\documentclass[letterpaper, 10 pt, conference]{ieeeconf}  
\IEEEoverridecommandlockouts                              

\usepackage{url}
\usepackage{graphics} 
\usepackage{epsfig} 
\usepackage{times} 
\usepackage{amsmath} 
\usepackage{amssymb}  
\usepackage{cite}
\usepackage{siunitx} 
\usepackage{soul}
\usepackage{color}
\usepackage{xcolor}

\soulregister\ref7  
\soulregister\cite7 
\soulregister\eqref7
\soulregister\prettyref7

\usepackage{multirow}
\usepackage{algpseudocode}
\usepackage{makecell}
\usepackage{booktabs}
\usepackage{threeparttable}
\usepackage[linesnumbered,ruled,vlined]{algorithm2e}
\usepackage{tabularx}

\bstctlcite{IEEEexample:BSTcontrol}

\usepackage{prettyref}
\newrefformat{fig}{Fig.~\ref{#1}}
\newrefformat{sec}{Sec.~\ref{#1}}
\newrefformat{tab}{Tab.~\ref{#1}}
\newrefformat{algo}{Algo.~\ref{#1}}
\newrefformat{eq}{(\ref{#1})}
\newrefformat{alli}{$line$~\ref{#1}}
\usepackage{arydshln}

\newcommand{\commentTom}[1]{\textcolor{black}{#1}}
\newcommand{\commentBenji}[1]{\textcolor{black}{#1}}

\newcommand\T{{\hspace{-0pt}\intercal}}
\title{\LARGE \bf
BagIt! An Adaptive Dual-Arm Manipulation of Fabric Bags for Object Bagging}
\author{Peng Zhou$^{1}$, Jiaming Qi$^{2}$, Hongmin Wu$^{3}$, Chen Wang$^{4}$, Yizhou Chen$^{4}$, Zeqing Zhang$^{4}$
\thanks{$^{1}$ The Great Bay University, Guangdong, China.}
\thanks{$^{2}$ Northeast Forestry University, Heilongjiang, China.}
\thanks{$^{3}$ Guangdong Academy of Sciences, Guangdong, China.}
\thanks{$^{4}$ The University of Hong Kong, Hong Kong, China.}
}

\begin{document}
\maketitle
\thispagestyle{empty}
\pagestyle{empty}

\begin{abstract}
Bagging tasks, commonly found in industrial scenarios, are challenging considering deformable bags' complicated and unpredictable nature. This paper presents an automated bagging system from the proposed adaptive Structure-of-Interest (SOI) manipulation strategy for dual robot arms. The system dynamically adjusts its actions based on real-time visual feedback, removing the need for pre-existing knowledge of bag properties. Our framework incorporates Gaussian Mixture Models (GMM) for estimating SOI states, optimization techniques for SOI generation, motion planning via Constrained Bidirectional Rapidly-exploring Random Tree (CBiRRT), and dual-arm coordination using Model Predictive Control (MPC). Extensive experiments validate the capability of our system to perform precise and robust bagging across various objects, showcasing its adaptability. 
This work offers a new solution for robotic deformable object manipulation (DOM), particularly in automated bagging tasks. Video of this work is available at \url{https://youtu.be/6JWjCOeTGiQ}.
\end{abstract}

\section{Introduction}
\label{sec:intro}
DOM involves the robotic handling of flexible objects that can change shape, presenting unique challenges compared to rigid object manipulation.
The primary challenges in DOM include the need for precise control, adaptability to varying material properties, and real-time responsiveness to complicated changes in the fabric state.
Research in DOM has progressed through the manipulation of 1D objects, such as ropes, where techniques focus on tension and knot formation; 2D objects, like garments, which emphasize grasping and folding strategies; and 3D objects, such as plush toys, requiring coordination and gentle handling. 
Among these, the bagging task in 3D DOM introduces additional complexities, as it involves enveloping objects with a deformable bag.
This task demands sophisticated planning and control strategies to ensure effective manipulation, accounting for the bag's flexibility and the dynamic constraints posed by the contents.

\begin{figure}
\centering
\includegraphics[width=0.44\textwidth]{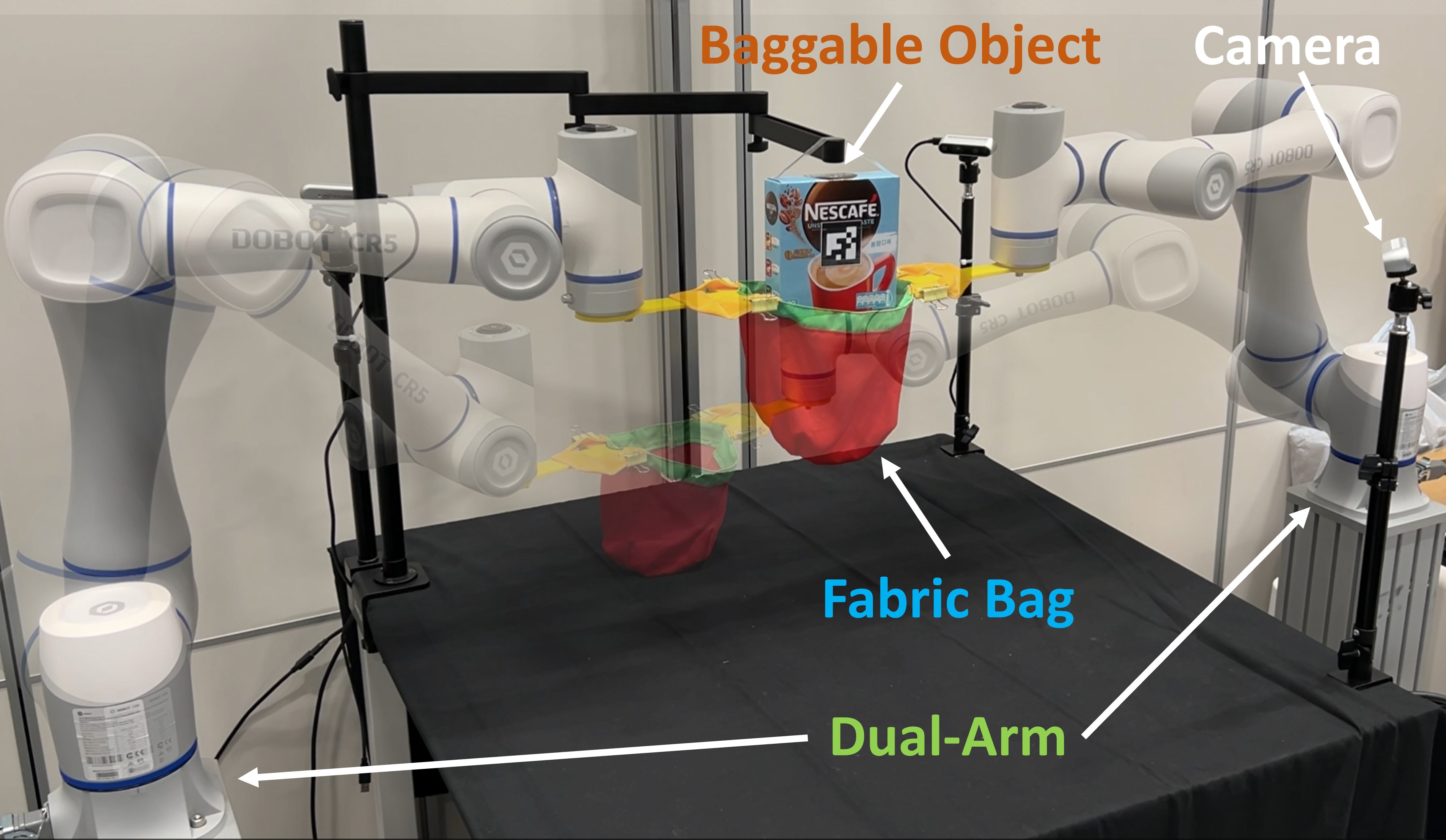}
\vspace{-10pt}
\caption{The dual-arm system grasps the handles of a fabric bag to control the SOI for the bagging operation.}
\label{fig:first_fig}
\vspace{-20pt}
\end{figure}

To this end, we employ a constraint-aware SOI planning framework to envelop 3D objects with a deformable fabric bag, as shown in \prettyref{fig:first_fig}. 
Our method draws on the concept of the Region of Interest (ROI) in image processing, indicating that fully estimating the state of a manipulated deformable object is not crucial in DOM.
Specifically, in the bagging task, the edge of a fabric bag acts as the SOI. By focusing on the state estimation of this edge, the robotic system successfully carries out the bagging operation.
Utilizing dual-arm operations, we represent the SOI as ellipses and introduce a two-stage manipulation strategy based on the object's bottom shape. An MPC-based controller ensures both arms accurately follow the planned trajectory. 
Our experiments demonstrate that in our framework, it is adequate to employ the SOI extracted merely from the bag opening rim to represent the whole fabric states during the bagging task. 
Furthermore, the presented system employs two robot arms working together, directed by an advanced planning framework that considers the object's structural limitations and the intended final configuration. Utilizing 3D-printed connectors, the robots achieve exceptional precision and stability in manipulating the bag. 

\noindent{\textbf{Main contributions}}: 

\begin{itemize}
    \item \commentBenji{We propose a robotic framework for automated bagging tasks, covering SOI estimation, bagging SOI generation, SOI planning, and a local planner for a dual-arm system.}
    \item \commentBenji{We introduce a constraint-aware planning strategy for SOI, enabling two robotic arms to successfully handle bags over various objects and achieve desired bag states.}
    \item \commentBenji{We integrate a vision-based control system that operates without prior knowledge of bag material properties, improving flexibility and adaptability in practice.}
\end{itemize}


\section{Related work}
\label{sec:related}

\commentBenji{DOM integrates perception, modeling, and planning to enable robots to handle soft, flexible materials \cite{gu2023survey}. Here, deformable objects have a broad scope in a generalized sense, encompassing granular materials \cite{zhang2025joint}, liquids \cite{niu2023goats}, and any non-rigid materials. In a narrow sense, DOM focuses more on robotic manipulation of ropes \cite{lv2023learning}, belts \cite{qi2026llm}, clothes \cite{yang2024one}, sponge \cite{navarro2017fourier}, etc.}
\commentBenji{Early DOM studies, e.g., \cite{yuba2014unfolding}, focus on single-arm systems that combine visual and force sensing to estimate deformation models for local control of fabrics and elastic objects. With advances in multimodal sensing and deep learning, researchers \cite{lips2024learning, yang2024one, shen2024action} have developed high-fidelity shape representations—from key point sets and point-cloud reconstructions to implicit neural fields—that support accurate global shape prediction and motion planning for complex deformable bodies. Concurrently, fast simulation \cite{lin2021softgym} and reinforcement learning \cite{niu2023goats} have yielded strategies that enable robots to adapt dynamically and execute multi-step manipulation despite uncertain initial configurations.}
\commentBenji{Dual-arm robotic systems extend the capabilities of single manipulators by providing additional degrees of freedom and spatial coordination. Recent works leverage two-arm coordination for tasks such as cloth folding \cite{zhang2023visual} and unfolding \cite{ha2022flingbot}, using symmetric grasping and multimodal perception. These approaches demonstrate improved handling precision but often target relatively simple deformable objects.}

\commentBenji{Despite these advances, the specific challenge of coordinating dual manipulators to handle deformable bags remains underexplored. Preliminary efforts \cite{gu2024shakingbot, chen2023autobag} utilize prior action primitives like shaking, integrated with visual guidance, to open and align bag mouths. 
However, this approach results in significant time consumption for each bagging operation. 
Recent work \cite{zhou2024bimanual} employs a graph neural network to represent and predict bag states, enabling efficient alignment of the bag rim to the target state. However, achieving a robust bagging task—where objects are placed reliably into a soft bag—introduces further challenges, including the goal state determination of the bag for the object insertion, maintaining consistent tension, and ensuring precise object insertion without excessive deformation.}
\commentBenji{To this end, we propose the concept of SOI as the bag state representation and introduce a constraint-aware SOI planning strategy for dual manipulators. Leveraging a visual-servoing controller, our method has been validated to robustly and efficiently complete various bagging tasks.}


\begin{figure}
\centering
\includegraphics[width=0.465\textwidth]{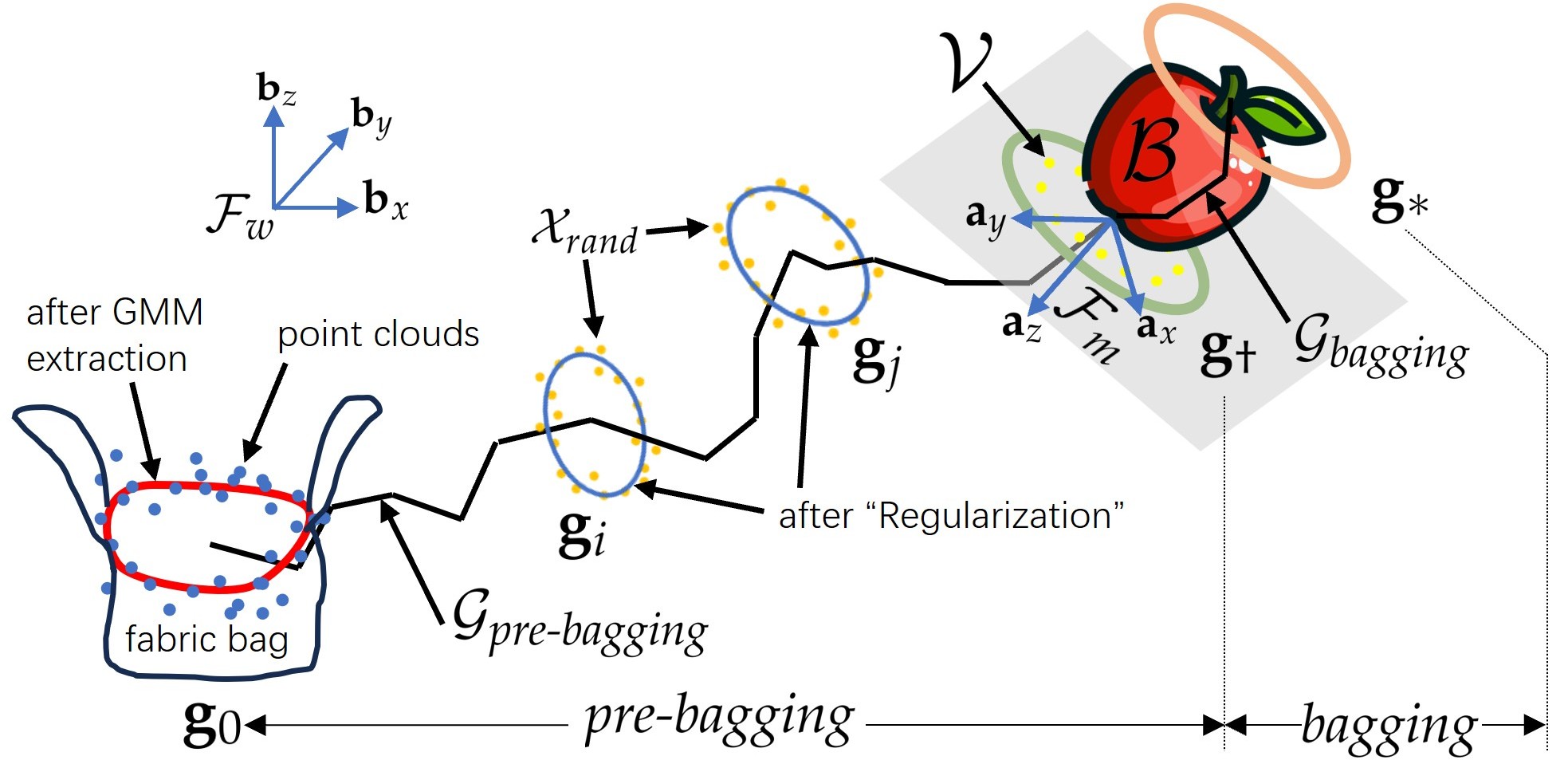}
\vspace{-10pt}
\caption{\commentBenji{Sketch of the proposed two-stage bagging strategy.}
}
\label{fig:bag_problem}
\vspace{-20pt}
\end{figure}

\section{Problem Statement}
\label{sec:prob_state}

The goal of the bagging task is to manipulate the handles of a fabric bag and wrap it around the bottom of the object $\mathcal{B}$, as shown in \prettyref{fig:bag_problem}. The bottom of $\mathcal{B}$ is represented by a set of vertices $\mathcal{V} = \{\mathbf{v}_i\}_{i=1}^{n_v}$. We assume that all points in $\mathcal{V}$ lie on the same plane, e.g., the gray surface under the apple in \prettyref{fig:bag_problem}. 
\commentBenji{We consider that the bag's initial shape should remain slightly open rather than fully closed. Furthermore, we assume that all bags involved in this task are of a type that humans can manage in some way.}

In this paper, we propose that a comprehensive estimation of the entire fabric bag is not requisite for the task at hand. Rather, our approach emphasizes the real-time identification and analysis of the SOI, which, in this context, is delineated as the opening rim of the bag.
The set composed of points that make up SOI at time $t$ is denoted by $\mathcal{Q}_t = \{\mathbf{q}_{i,t}\}_{i=1}^{n_x}$.
Furthermore, the $\mathcal{Q}_t$ can be extracted from the raw point clouds $\mathcal{P}_t = \{\mathbf{p}_{i,t}\}_{i=1}^{n_p}$ via our proposed extraction method (Sec. \ref{section4a}). For example, in \prettyref{fig:bag_problem}, in the beginning, blue dots refer to point clouds $\mathcal{P}_0$, and the red curve is formed by points of SOI $\mathcal{Q}_0$. 
Here $\mathbf{v}_i, \mathbf{q}_{i,t}, \mathbf{p}_{i,t} \in \mathbb{R}^3$ are all 3D point expressed in the world frame $\mathcal{F}_w$, as shown in \prettyref{fig:bag_problem}. And $^{m}\mathbf{q}_{i,t}$ indicates the point  $\mathbf{q}_{i,t}$ expressed in a frame $\mathcal{F}_m$. Note that unless otherwise noted, all points are expressed in $\mathcal{F}_w$ and will be omitted superscript $^{w}(\cdot)$ for simplicity.
In this way, the bag state $\mathbf{x}_t$ is defined by its SOI points, i.e.,:
\begin{align}\label{eq:soi_def}
    \mathbf{x}_t = \left[ \mathbf{q}_{1,t}^\intercal, \ldots, \mathbf{q}_{n_x,t}^\intercal \right]^\intercal
    \in \mathbb{R}^{3n_x}.
\end{align}
So, in the following, the bag state is equivalent to the SOI.

We partition the bagging task into two distinct phases based on whether the target object starts to enter the bag, as depicted in \prettyref{fig:bag_problem}. This critical criterion is defined as the bagging SOI, denoted by $\mathbf{g}_\dag$ (Sec.~\ref{section4b}). 
During the packing process, $\mathbf{g}_\dag$ naturally needs to satisfy two requirements. First, the size of the bag opening must be larger than the maximum cross-section of the object (i.e., size encircled by  $\mathcal{V}$) to ensure that the object can smoothly pass through the opening and enter the bag. Second, the size of the bag opening cannot be arbitrarily large due to the physical constraints of the actual bag. Therefore, we assume that the size of the bag opening remains relatively unchanged during the bagging process, staying close to its initial size (i.e., encircled size of $\mathcal{Q}_0$) while satisfying the first condition.

From $\mathbf{g}_\dag$, the goal SOI $\mathbf{g}_*$, where the bag fully warps the object, could be determined accordingly (\prettyref{sec:goal_soi}). Intuitively, the bag's initial SOI is given as $\mathbf{g}_0:= \mathbf{x}_0$, which can be specified manually before bagging. 
Based on $(\mathbf{g}_0,\mathbf{g}_\dag,\mathbf{g}_*)$, the bagging process can be separated as \textit{pre-bagging} and \textit{bagging} stages, respectively. Then we employ a sampling-based motion planner (see \prettyref{sec:soi_plan}) to generate subgoals $\mathbf{g}_i$ in each stage, as shown in \prettyref{fig:bag_problem}, and the collision-free bagging path can be formulated by:
\begin{align}\label{eq:bagging_path}
    \mathcal{G} := \{ \mathcal{G}_{\textit{pre-bagging}}, \mathcal{G}_{\textit{bagging}} \} = \{ \underbrace{\mathbf{g}_0, \mathbf{g}_1, \ldots,}_{\text{pre-bagging}} \underbrace{\mathbf{g}_\dag, \ldots, \mathbf{g}_*}_{\text{bagging}}\}.
\end{align}
Given the planned path $\mathcal{G}$, an MPC-based shape servoing method, as introduced in \prettyref{sec:motion_plan}, is utilized to generate the velocity commands $\mathbf{u}_t$ from the robotic action space $\mathcal{A}$ based on the current SOI state $\mathbf{x}_t$ and the corresponding next subgoal $\mathbf{g}_{next} \in \mathcal{G}$, i.e.,
\begin{align}\label{eq:mpc_opt}
    \mathbf{u}_t = \underset{\mathbf{u}_t \in \mathcal{A}}{\arg\min}~\mathbb{E} (\mathbf{x}_t, \mathbf{g}_{next}),
\end{align}
where $\mathbb{E}$ is a measurable error function. 
\commentTom{
The dynamic model between $\mathbf{x}_t$ and $\mathbf{u}_t$ will be discussed later in \eqref{eq:sys_model}.
}
Thus, the bagging task can be completed via a sequence of command velocities $\mathcal{U} = \{\mathbf{u}_t\}_{t=0}^T$ until $\mathbf{x}_t$ meets the goal SOI $\mathbf{g}_*$.

\begin{figure*}
\centering
\includegraphics[width=0.90\textwidth]{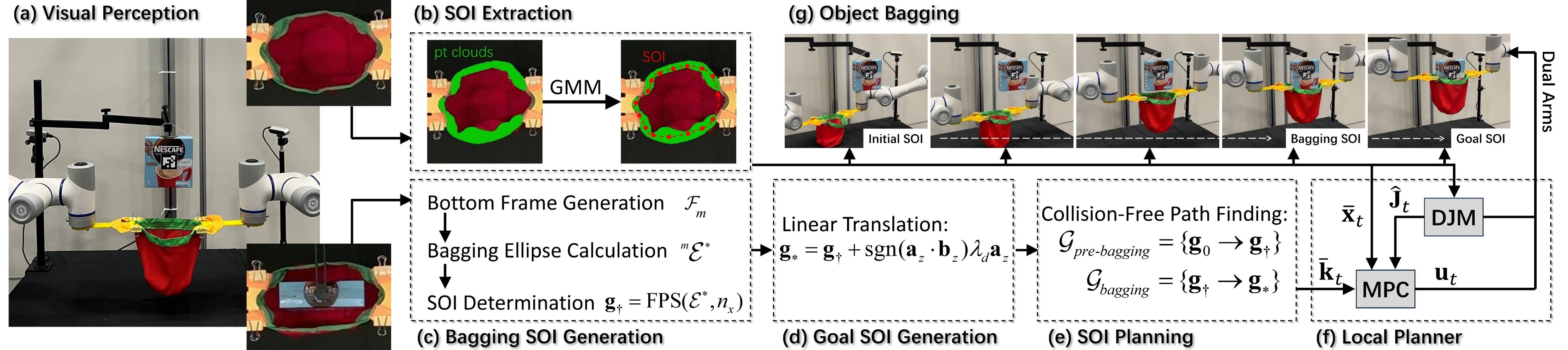}
\vspace{-10pt}
\caption{\commentBenji{Flowchart of our dual-arm manipulation strategy for the bagging task.}}
\label{fig:flowchart}
\vspace{-20pt}
\end{figure*}


\section{\commentTom{Visual Perception}}
\label{section4a}
Following the notions in \prettyref{sec:prob_state}, the SOI $\mathbf{x}_t$ is composed of $n_x$ contour points in $\mathcal{Q}_t$, capturing the bag's opening rim. 
In practice, the opening boundary of the bag is obtained via the depth camera in the form of raw point clouds $\mathcal{P}_t$ with $n_p$ points, as given in \prettyref{fig:flowchart}-(a). Normally $n_p \gg n_x$. This subsection presents a GMM-based SOI Extractor, a module for real-time SOI  $\mathcal{Q}_t$ extraction from dense, noisy point clouds $\mathcal{P}_t$, as shown in  \prettyref{fig:flowchart}-(b).

We assume all points in $\mathcal{P}_t$ are sampled from a mixture of $n_x$ Gaussian distributions with unknown parameters (i.e., means, covariances, and weights). 
To solve these parameters of $n_x$ Gaussian distributions, we will extract the $\mathcal{Q}_t$. 
This process can be treated as a probability density estimation problem using GMM.
Here, 
$\mathcal{P}_t$ is treated as points randomly sampled from the GMM, with the means of the Gaussians represented as 
$\mathcal{Q}_t$. 
To address the challenges posed by the outlier-prone nature of $\mathcal{P}_t$, a uniform distribution is incorporated into the GMM for $\mathcal{Q}_t$, enhancing robustness and accuracy in the extraction process.
The probability of $\mathbf{p}_{i,t}$ ($i=1,...,n_p$) to be sampled can be formulated as:
\begin{equation}
\label{eq48}
p (\mathbf{p}_{i,t}) 
= \sum_{\mathcal{Q}_t} p (\mathcal{Q}_t)   p (\mathbf{p}_{i,t} | \mathcal{Q}_t) 
= \sum_{j=1}^{n_x}\pi_j \mathcal{N}(\mathbf{p}_{i,t}|{\mu}_j,\Sigma_j) .
\end{equation}
Here $\pi_j$, $\mu_j$ and $\Sigma_j$ refer to the weight, mean, and covariance of $j$-th Gaussian distribution $\mathcal{N}$, respectively. Note that $\sum \pi_j = 1$. 
To generate $\mathbf{q}_{j,t}$ to form the bag rim $\mathcal{Q}_t$, we need to estimate parameters $\mu_j$ for all $n_x$ Gaussian distributions, since $\mu_j \in \mathbb{R}^3$ indicates the center position of each $\mathbf{q}_{j,t}$. 
To this end, the Expectation-Maximization (EM) algorithm can be employed to maximize the log-likelihood of all points in $\mathcal{P}_t$. 
%
As shown in \prettyref{fig:flowchart}-(b), the dense green dots represent $\mathcal{P}_t$, while the red dots $\mathcal{Q}_t$ extracted using GMM, which are subsequently identified as the bag state $\mathbf{x}_t$ by \prettyref{eq:soi_def}.


\section{\commentTom{SOI Generation}}\label{sec:method}

\subsection{Bagging SOI Generation}
\label{section4b}

As mentioned in \prettyref{sec:prob_state}, the bagging SOI $\mathbf{g}_\dag$, marking the entry of the object into the fabric bag, is crucial. In this subsection, we introduce the bagging SOI generation algorithm. As illustrated in \prettyref{fig:flowchart}-(c), it consists of three main steps, detailed in the following three subsubsections.

\subsubsection{Bottom Frame Creation}
\label{sec:fm_gen}

First, we create a local frame $\mathcal{F}_m$ positioned at the bottom of the baggable object $\mathcal{B}$. This allows the bag to envelop the object from below when the bagging SOI $\mathbf{g}_\dag$ is determined on the $xy$-plane of $\mathcal{F}_m$.

From \prettyref{sec:prob_state}, we can obtain a set of coplanar vertices $\mathcal{V} = \{\mathbf{v}_i\}_{i = 1}^{n_v}$ at the bottom of $\mathcal{B}$, as shown as yellow dots in  \prettyref{fig:bag_problem}. Thus, we define the plane containing $\mathcal{V}$ as the 
$xy$-plane of $\mathcal{F}_m$, as illustrated as a gray plane under the apple in \prettyref{fig:bag_problem}. If denoting the centroid of $\mathcal{V}$ as $\bar{\mathbf{v}} = \frac{1}{n_v} \Sigma \mathbf{v}_i$, the $x,y,z$-axis of $\mathcal{F}_m$ can be determined from the following equations, accordingly.
\begin{align}\label{eq:xyz_fm}
    \mathbf{a}_z = \frac{\Delta \mathbf{v}_1\times \Delta \mathbf{v}_2}{\|\Delta \mathbf{v}_1\times \Delta \mathbf{v}_2\|}, \mathbf{a}_y = \frac{\Delta \mathbf{v}_3}{\|\Delta \mathbf{v}_3\|},
    \mathbf{a}_x = \frac{\mathbf{a}_y \times \mathbf{a}_z}{\|\mathbf{a}_y \times \mathbf{a}_z\|}
\end{align}
where $\Delta \mathbf{v}_k = \mathbf{v}_k - \bar{\mathbf{v}}$, ($k=1,2,3$) and $ \mathbf{v}_1,  \mathbf{v}_2, \mathbf{v}_3$ are randomly selected from $\mathcal{V}$. Intuitively, the $\bar{\mathbf{v}}$ is selected as the origin of $\mathcal{F}_m$. Then, the transformation matrix {from bottom frame $\mathcal{F}_m$ to world frame $\mathcal{F}_w$}  can be expressed as:
\begin{align}\label{eq:transf_matrix}
^{w} \mathbf{T}_{m} 
= 
\left[
\begin{array}{cccc}
{\mathbf{a}_x} & {\mathbf{a}_y} & {\mathbf{a}_z} & {\bar{\mathbf{v}}}  \\
0 & 0 & 0 & 1 
\end{array}
\right] 
\in \mathbb{R}^{4 \times 4}.
\end{align}
As such, given the $\mathcal{V} = \{\mathbf{v}_i\}_{i = 1}^{n_v}$ expressed in  $\mathcal{F}_w$, the new coordinates in the $\mathcal{F}_m$ can be calculated by
$^m\mathcal{V} = \{
    ^m\mathbf{v}_i := [{}^mv_{i,x}, {}^mv_{i,y}, {}^mv_{i,z}]^\intercal|
    [{\mathbf{a}_x}, {\mathbf{a}_y}, {\mathbf{a}_z}]^\intercal (\mathbf{v}_i - \mathbf{\bar{\mathbf{v}}}), \forall \mathbf{v}_i \in \mathcal{V} \}_{i = 1}^{n_v}$.
Since $\mathbf{v}_i$ are coplanar, then ${}^mv_{i,z} \equiv 0$ and ${}^m\bar{\mathbf{v}} \equiv \mathbf{0}$.


\subsubsection{Bagging Ellipse Calculation}
\label{sec:bag_ellipse}
Given the symmetrical manipulation of the bag by dual arms, the bagging SOI $\mathbf{g}_\dag$ will be formulated as an ellipse shape. 
To determine this ellipse, we formulate it as an optimization problem considering the minimum of perimeters between the ellipse and the bag rim, that is,
\begin{equation}\label{eq:ellipse_opti}
\begin{aligned}
    &{\mathop{\arg\min}_{\commentTom{^m\tau_x,^m\tau_y,\rho_a,\rho_b,\alpha}}}  \| L - R \|^2\\
    \text{s.t.}~ C_1~\text{from}~&\prettyref{eq:constraint_1},~ 
    C_2~\text{from}~\prettyref{eq:constraint_2},~
    C_3~\text{from}~\prettyref{eq:constraint_3}.
    \end{aligned}
\end{equation}
for $L$ is the ellipse's perimeter and $R$ is the perimeter of the observed bag's opening. 
As assumed in \prettyref{sec:prob_state}, the bag opening is assumed to remain relatively stable during the bagging process. 
$R$ can be derived from the initial rim observation $\mathcal{Q}_0$, i.e., $R = \sum \|\mathbf{q}_{i,0} - \mathbf{q}_{i+1,0}\|$.

{\textbf{Ellipse Formulation}}
From analytic geometry, the parametric equations for a 2D ellipse centered at $(^m\tau_x,^m\tau_y)$ orientated at $\alpha$ degree is written as:
\begin{equation}\label{eq:ellipse_2d_para}
    \begin{aligned}
    x &= {}^m\tau_x + \rho_a  \cos(\theta)  \cos(\alpha) - \rho_b \sin(\theta) \sin(\alpha), \\
    y &= {}^m\tau_y + \rho_a  \cos(\theta) \sin(\alpha) + \rho_b \sin(\theta) \cos(\alpha),
\end{aligned}
\end{equation}
where $\theta \in [0, 2\pi]$ and $\rho_a,\rho_b$ are semi-major and semi-minor axes, respectively.
Here $(^m\tau_x,^m\tau_y)$ indicates this ellipse is located on the $xy$-plane of $\mathcal{F}_m$. 
\commentTom{
We use the method in \cite{abramowitz1948handbook} to calculate the perimeter $L$ of the parametric ellipse \eqref{eq:ellipse_2d_para}.
}


{\textbf{Constraint $C_1$: Limit the size of the ellipse to be larger than the bottom of the object}.}
Base on \prettyref{eq:ellipse_2d_para}, if a point $(x, y)$ is located in the ellipse, it satisfies $\mathbb{F}_e(x,y) < 1$,
where $\mathbb{F}_e = {((x-{}^m\tau_x)\cos{\alpha} + (y-{}^m\tau_y)\sin{\alpha})^2}/{\rho^2_a} + {((x-{}^m\tau_x)\sin{\alpha} + (y-{}^m\tau_y)\cos{\alpha})^2}/{\rho^2_b}$. 
So, we can formulate the constraint $C_1$ as follows:
\begin{align}\label{eq:constraint_1}
    0 \leq \mathbb{F}_e({}^m\nu) < \lambda_1,
\end{align}
where ${}^m\nu$ contains $xy$-coordinates of all $^m\mathbf{v}_i\in {}^m\mathcal{V}$, as defined by $^m\nu:= \{({}^mv_{i,x}, {}^mv_{i,y}) |~{}^m\mathbf{v}_i \in {}^m\mathcal{V}\}$, and $\lambda_1$ controls the extent to which the ellipse envelops the base of the object. To make sure the bottom of $\mathcal{B}$ is inside of the ellipse, 
$\lambda_1 < 1$ to satisfy $\mathbb{F}_e(x,y) < 1$.

\textbf{Constraint $C_2$: Limit concentricity of the ellipse and the base of object.}

The concentricity between the center of ellipse and the centroid of object base can be evaluated by their Euclidean distance, thus we yield:
$
    \|[{}^m\tau_x, {}^m\tau_y, 0]^\intercal - {}^m\bar{\mathbf{v}} \|\leq \lambda_2. \nonumber
$
From \prettyref{sec:fm_gen}, due to the coplanarity of $\mathcal{V}$, it has ${}^m\bar{\mathbf{v}} \equiv \mathbf{0}$. So, we generate the constraint $C_2$ as
\begin{align}\label{eq:constraint_2}
    \sqrt{{}^m\tau_x^2 + {}^m\tau_y^2} \leq \lambda_2.
\end{align}

{\textbf{Constraint $C_3$: Limit the ellipse orientation according to object base orientation.}}
We restrict the orientation of the ellipse based on the parallelism between the major axis of the ellipse (denoted by $\mathbf{d}_e \in \mathbb{R}^2$) and the principal axis of the object base vertices $^m\mathcal{V}$ (denoted by $\mathbf{d}_v \in \mathbb{R}^2$). 
If we discretize $\theta$ uniformly from $[0, 2\pi]$, the set of points on the ellipse \prettyref{eq:ellipse_2d_para} is obtained as $^m\mathcal{E} = \{(x_i, y_i) | \prettyref{eq:ellipse_2d_para} \leftarrow \theta_i \sim \mathcal{U}(0,2\pi) \}_{i = 1}^{n_e} $. 
To determine the principal axis direction, the principal components analysis (PCA) is utilized, such that $\mathbf{d}_e = \text{PCA}(^m\mathcal{E})$ and $\mathbf{d}_v = \text{PCA}(^m\mathcal{V})$. 
So, the degree of parallelism can be measured by the inner dot product of two principal axes, given by the constraint $C_3$:
\begin{align}\label{eq:constraint_3}
    \lambda_3 \leq | \mathbf{\mathbf{d}}_e \cdot \mathbf{d}_v | \leq 1.
\end{align}
%
Since $\mathbf{d}_e$ and $\mathbf{d}_v$ are unit vectors from PCA, $\mathbf{d}_e \cdot \mathbf{d}_v$ represents the cosine of the angle between them, ranging from $-1$ to $1$. We focus on parallelism, so both the same direction (i.e., $\mathbf{d}_e \cdot \mathbf{d}_v = 1$) and opposite direction (i.e., $\mathbf{d}_e \cdot \mathbf{d}_v = -1$) are desired. Thus, the absolute operation is specified in \prettyref{eq:constraint_3}.

\commentTom{
Specifically, $C_2$ and $C_3$ act as adjustment factors for $C_1$, helping to find a suitable ellipse. 
Without $C_1$, there is a risk of collision between the ellipse's edge and the object. 
Including $C_2$ and $C_3$ centers the ellipse and moves the bottom vertices away from the edge, reducing collision risk.
}
By solving the optimization problem \prettyref{eq:ellipse_opti}, the bagging ellipse is obtained, i.e., green ellipse at the bottom of the apple in \prettyref{fig:bag_problem} denoted as ${}^m \mathcal{E}^*$.
In the next, we will demonstrate the bagging SOI $\mathbf{g}_\dag$ generation based on this ${}^m \mathcal{E}^*$.


\subsubsection{SOI Determination}
\label{sec:downsampling}
At first, the coordinates of $^m\mathcal{E}^*$ should be transformed in the world frame $\mathcal{F}_w$. It can be done by \prettyref{eq:transf_matrix} via $\mathcal{E}^* = ^{w} \mathbf{T}_{m}  {}^m\mathcal{E}^*$. Also, we can find that there are $n_x$ points needed to generate SOI from \prettyref{eq:soi_def}, while $n_e$ points exist in the $\mathcal{E}^*$. Normally, $n_e \gg n_x$. So, next, a downsampling operation while maintaining the shape of the resulting ellipse is required. We adopt the farthest point sampling (FPS) method \cite{yan2020pointasnl} to yield $n_x$ points from $\mathcal{E}^*$:
\begin{align}\label{eq:goal_soi_gen}
    \mathcal{Q}^* = \text{FPS}(\mathcal{E}^*, n_x).
\end{align}
Thus, the bagging SOI can be generated as $\mathbf{g}_\dag := \mathbf{x}_\dag$ by substituting $\mathcal{Q}^*$ into \prettyref{eq:soi_def}, as displayed in \prettyref{fig:bag_problem}.

\subsection{Goal SOI Generation}
\label{sec:goal_soi}
According to \prettyref{fig:flowchart}-(d), in this subsection, we will calculate the goal SOI $\mathbf{g}_*$.
As defined in \prettyref{eq:bagging_path}, it can be seen that $\mathbf{g}_\dag$ is a critical state between the \textit{pre-packaging} and \textit{packaging} processes. After $\mathbf{g}_\dag$, the object begins to enter the fabric bag, so the final bag state can be determined by the simple translation of $\mathbf{g}_\dag$ in the direction toward the object, as shown as the orange ellipse in \prettyref{fig:bag_problem}. 
Mathematically, the goal SOI $\mathbf{g}_*$ is formulated by
\begin{align}\label{eq:g_star_def}
\mathbf{g}_* = \mathbf{g}_\dag + {\rm{sgn}}(\mathbf{a}_z \cdot \mathbf{b}_z) \lambda_d \mathbf{a}_z,
\end{align}
for $\lambda_d > 0$, adjusting the depth of the object into the bag.
$\mathbf{a}_z$ is the $z$-axis of $\mathcal{F}_m$ from \prettyref{eq:xyz_fm}.
$\mathbf{b}_z=[0,0,1]^\T$ is the $z$-axis of $\mathcal{F}_w$.
Principally, $\mathbf{g}_{\ast}$ is generated by extending $\mathbf{g}_{\dag}$ upwards along the side of $\mathcal{B}$.
However, as $\mathbf{v}_1$ and $\mathbf{v}_2$ are arbitrarily selected, $\mathbf{a}_z$ cannot be determined as either inward or outward the object. 
Thus, the sign function $\rm{sgn}(\cdot)$ is utilized to invert $\mathbf{a}_z$ if $\mathbf{a}_z$ is outward.
E.g. in the case of \prettyref{fig:bag_problem}, $\rm{sgn}(\cdot) = -1$, as $\mathbf{a}_z$ points to the outside of the object.


\vspace{-12pt}
\begin{algorithm}[!htbp]  
    \footnotesize
    \caption{SOI Planning: $\mathbf{g}_0 \rightarrow \mathbf{g}_\dag$}
    \label{algo:soi_path_plan}
    \SetKwInOut{Input}{Input}\SetKwInOut{Output}{Output}
    \Input{Initial SOI $\mathbf{g}_0$, Bagging SOI $\mathbf{g}_\dag$} 
    \Output{Path $\mathcal{G}_\textit{pre-bagging}$} 
    Initialize two trees: $T_{\text{start}}$ and $T_{\text{goal}}$\label{alli:init_s}\\
    Add $\mathbf{g}_0$ to $T_{\text{start}}$, $\mathbf{g}_\dag$ to $T_{\text{goal}}$\label{alli:init_e}\\
    \For{$i = 1$ \KwTo $\text{MAX}_{iteration}$\label{alli:iter_max}}{
        $\mathbf{x}_{rand}$ = SampleRandomPoint()\label{alli:sample_x}\\
        $\mathbf{x}^e_{rand}$ = Regularization($\mathbf{x}_{rand}$)\label{alli:project_stable}\\
        $\mathbf{x}_{nearest\_start}$ = NearestNeighbor($T_{\text{start}}$, $\mathbf{x}^e_{rand}$)\label{alli:extend1}\\
        $\mathbf{x}_{new\_start}$ = Extend($\mathbf{x}_{nearest\_start}$, $\mathbf{x}^e_{rand}$, constraints)\label{alli:extend2}\\

        \If{$\mathbf{x}_{new\_start}$ is not null}{
            $\mathbf{g}_{new\_start} \leftarrow \mathbf{x}_{new\_start}$ \label{alli:subgoal1}\\
            Add $\mathbf{g}_{new\_start}$ to $T_{\text{start}}$ \label{alli:add_tree1}\\
            $\mathbf{x}_{nearest\_goal}$ = NearestNeighbor($T_{\text{goal}}$, $\mathbf{x}_{new\_start}$)\label{alli:check_proxi1}\\
            \If{Distance($\mathbf{x}_{new\_start}$, $\mathbf{x}_{nearest\_goal}$) $< \varepsilon$\label{alli:close_enough_s1}}{
                Connect($\mathbf{x}_{new\_start}$, $\mathbf{x}_{nearest\_goal}$)\\
                {\bf return} Path $\mathcal{G}_\textit{pre-bagging}$ from start to goal \label{alli:close_enough_e1}\\
            }
        }   
        $\mathbf{x}_{nearest\_goal}$ = NearestNeighbor($T_{\text{goal}}$, $\mathbf{x}^e_{rand}$)\label{alli:extend3}\\
        $\mathbf{x}_{new\_goal}$ = Extend($\mathbf{x}_{nearest\_goal}$, $\mathbf{x}^e_{rand}$, constraints)\label{alli:extend4}\\
        \If{$\mathbf{x}_{new\_goal}$ is not null}{
            $\mathbf{g}_{new\_goal} \leftarrow \mathbf{x}_{new\_goal}$\label{alli:subgoal2}\\
            Add $\mathbf{g}_{new\_goal}$ to $T_{\text{goal}}$\label{alli:add_tree2}\\
            $\mathbf{x}_{nearest\_start}$ = NearestNeighbor($T_{\text{start}}$, $\mathbf{x}_{new\_goal}$)\label{alli:check_proxi2}\\
            \If{Distance($\mathbf{x}_{new\_goal}$, $\mathbf{x}_{nearest\_start}$) $< \varepsilon$\label{alli:close_enough_s2}}{
                Connect($\mathbf{x}_{new\_goal}$, $\mathbf{x}_{nearest\_start}$)\\
                {\bf return} Path $\mathcal{G}_\textit{pre-bagging}$ from start to goal \label{alli:close_enough_e2}\\
            }
        }
    }
    {\bf return} Failure \label{alli:failure}
\end{algorithm}
\vspace{-20pt}

\section{\commentTom{Manipulation Planning}}
\subsection{SOI Planning}
\label{sec:soi_plan}
In this section, we will show the path planning from $\mathbf{g}_0$ to $\mathbf{g}_\dag$ and from $\mathbf{g}_\dag$ to $\mathbf{g}_*$, respectively, as demonstrated in \prettyref{fig:flowchart}-(e).
Taking $\mathbf{g}_0 \rightarrow \mathbf{g}_\dag$ as an example, the constrained Bi-directional Rapidly-exploring Random Tree (CBiRRT) \cite{yu2023coarse} is employed in the $\mathcal{F}_w$ to find a sequence of subgoals, as summarized the  \prettyref{algo:soi_path_plan}. To improve the sampling feasibility, we add the ``Regularization'' function, used to regularize the randomly sampled state into a standard ellipse, as shown in \prettyref{alli:project_stable} of \prettyref{algo:soi_path_plan}.

\subsubsection{Sampling-based Path Finding}
CBiRRT is an advanced path planning algorithm used in high-dimensional spaces that must navigate around obstacles while adhering to specific constraints. From \prettyref{algo:soi_path_plan}, the algorithm begins by creating two trees: one rooted at the start point and the other at the goal point (\prettyref{alli:init_s}-\prettyref{alli:init_e}). It will simultaneously grow two trees, each exploring 3D space of $\mathcal{F}_w$ in search of a valid \commentBenji{and collision-free} path. In each iteration, a random point in the space is sampled (\prettyref{alli:sample_x}).
To improve the subsequent MPC-based shape servoing control over the bag rim, we expect the subgoals along the path follow an elliptical shape. 
However, it is evident that the $n_x$ points from random sampling $\mathbf{x}_{rand}$ hardly guarantee that the shape they form is (or is close to) an ellipse.  
To address this, we propose a regularization function ``Regularization'' which utilizes an optimization algorithm to approximate the shape formed by $\mathbf{x}_{rand}$ with a standard ellipse $\mathbf{x}^e_{rand}$, as in \prettyref{alli:project_stable}. After that, the algorithm attempts to extend both trees toward this random state (\prettyref{alli:extend1}, \prettyref{alli:extend3}) while ensuring that the growth respects any imposed constraints (\prettyref{alli:extend2}, \prettyref{alli:extend4}). 
The constraints here primarily consider the collision-free condition and the perimeter-consistency condition. If a new state is successfully added to one tree (\prettyref{alli:add_tree1}, \prettyref{alli:add_tree2}) as a subgoal along the path (\prettyref{alli:subgoal1}, \prettyref{alli:subgoal2}), the algorithm checks for proximity to nodes in the other tree (\prettyref{alli:check_proxi1}, \prettyref{alli:check_proxi2}). If they are close enough within a threshold $\varepsilon$, the trees are connected, yielding a feasible path (\prettyref{alli:close_enough_s1}-\prettyref{alli:close_enough_e1}, \prettyref{alli:close_enough_s2}-\prettyref{alli:close_enough_e2}). 
The process continues for a maximum number of iterations (\prettyref{alli:iter_max}) or until a valid path $\mathcal{G}_\textit{pre-bagging}$ is found. If no path is discovered, the algorithm reports failure (\prettyref{alli:failure}). Similarly, given $\mathbf{g}_\dag$ and $\mathbf{g}_*$, the valid path $\mathcal{G}_\textit{bagging}$ could also be found. Finally, the whole path $\mathcal{G}$ is formed in \prettyref{eq:bagging_path}.

\commentTom{
As the lack of accurate robot-bag relationship, the infinite dimensions of the bag in fixed arm setups may cause unexpected deformations when sampling relative poses in CBiRRT.
This paper builds a local robot-bag relationship, using shape servoing to convert shape control of the SOI into commands for robot motion, ensuring reliable manipulation.
}
\commentTom{
CBiRRT does not ensure optimality; it only finds a collision-free connection between $\mathbf{g}_0$ and $\mathbf{g}_{\ast}$.
By adding path trimming and local smoothing during planning, the path's compactness can be improved.
Incorporating shape constraints for the SOI during planning helps reduce unnecessary openings and closings of intermediate shapes.
}

\subsubsection{Regularization}
Given a set of points, we propose the ``Regularization'' function to estimate its shape outlined by those points using a standard ellipse with parameters $\eta$. This operation is also represented in \prettyref{fig:bag_problem} for subgoals $\mathbf{g}_i, \mathbf{g}_j$. We formulate it as an optimization problem:
\begin{equation}\label{eq:project_stable}
    \begin{aligned}
        &\underset{{\eta}}{\arg\min} \| \rm{CD}(\mathcal{Y},\mathcal{X}_{rand})\|^2 \\
        \text{s.t.}~ 
        &C_4~\text{from}~\prettyref{eq:constraint_4},
        C_5~\text{from}~\prettyref{eq:constraint_5}.
    \end{aligned}
\end{equation}
The $\rm{CD}(\cdot)$ above indicates the Chamfer Distance, to evaluate the similarity between two sets of points with different numbers. Here $\mathcal{X}_{rand}$ refers to a set of points from $\mathbf{x}_{rand}$ in \prettyref{alli:sample_x}, and $\mathcal{Y}$ represents the set of ellipse points, determined as follows. Given a standard equation for a 3D ellipse $\mathbb{E}$, 
\begin{align}\label{eq:ellip_eq}
    \mathbb{E}(\theta)|_{\eta = \{\mathbf{c}, \mathbf{u}, \mathbf{v}, \rho_u, \rho_v\}}  &= \mathbf{c} + \rho_u\cos({\theta}) \mathbf{u} + \rho_v \sin(\theta) \mathbf{v},
\end{align}
its parameters $\eta$ include center $\mathbf{c} \in \mathbb{R}^3$, semi-major(minor) axis vector $\mathbf{u} \in \mathbb{R}^3$($\mathbf{v} \in \mathbb{R}^3$) and its length $\rho_u$($\rho_v$).
We can generate $\mathcal{Y}$ via the uniformly discretizing the input $\theta$ from $[0,2\pi]$, i.e., 
$
    \mathcal{Y} = \{\mathbf{y}_i | \prettyref{eq:ellip_eq} \leftarrow \theta_i \sim \mathcal{U}(0,2\pi)\}_{i=1}^{n_y}.
$
Accordingly, the perimeter of this ellipse can be approximated by $\mathcal{R}_y = \sum \|\mathbf{y}_i - \mathbf{y}_{i+1}\|$. The perimeter of the shape outlined by all points in $\mathcal{X}_{rand}$ could be calculated similarly, denoted by  $\mathcal{R}_x$.
Constraints in \prettyref{eq:project_stable} are formulated below.

\textbf{Constraint $C_4$: Limit two perimeters to roughly the same.}
Given $\mathcal{R}_x$ and $\mathcal{R}_y$, the constraint $C_4$ is given as:
\begin{align}\label{eq:constraint_4}
     |{\mathcal{R}_x}/{\mathcal{R}_y} - 1| \leq \lambda_4.
\end{align}
The parameter $\lambda_4$ limits the scale between two perimeters.

\textbf{Constraint $C_5$: Limit concentricity of two shapes.}
Similar to the constraint $C_2$ in \prettyref{eq:ellipse_opti}, we want the ellipse's center $\mathbf{c}$ to coincide with the centroid of $\mathcal{X}_{rand}$ (denoted by $\bar{\mathbf{c}}$) as well. So, constraint $C_5$ can be expressed as:
\begin{align}\label{eq:constraint_5}
    \|\mathbf{c} - \bar{\mathbf{c}}\| \leq \lambda_5.
\end{align}
The parameter $\lambda_5$ restricts the distance between two centers.


By solving the optimal problem \prettyref{eq:project_stable}, we obtain an ellipse with parameters $\eta^*$, which regularizes the shape of the randomly sampled $\mathcal{X}_{rand}$, {as exemplified in  \prettyref{fig:bag_problem}.}


\subsection{\commentTom{
Local Planner for Manipulator Motion Generation
}}
\label{sec:motion_plan}

From \prettyref{fig:flowchart}-(f), we present \commentTom{a local planner} to ensure both arms manipulate the bag to follow the planned trajectory. 
We model the manipulation process using an MPC-based shape servoing approach, similar to \cite{qi2023adaptive}, where incremental robot movements cause small deformations in the bag. 
As the used dual-arm, the pose command is denoted by $\mathbf{r}\in\mathbb{R}^{12}$.
As in \cite{qi2021contour}, the system model can be defined by  
\begin{align}\label{eq:sys_model}
    \mathbf{s}_t = \mathbf{J}_t  \mathbf{u}_t, 
\end{align}
where $\mathbf{s}_t = \mathbf{x}_t - \mathbf{x}_{t-1} \in \mathbb{R}^{3n_x}$ (see \prettyref{eq:soi_def})$, \mathbf{u}_t = \mathbf{r}_t - \mathbf{r}_{t-1} \in \mathbb{R}^{12}$. The deformation Jacobian matrix (DJM) $\mathbf{J}_t \in \mathbb{R}^{3n_x \times 12}$ represents the relationship between the increment of dual-arm manipulation (i.e., $\mathbf{u}_t$) and changes in the bag's SOI (i.e., $\mathbf{s}_t$). In addition, 
we assume that $\mathbf{J}_t$ maintains full column rank through the task. This condition is generally straightforward to satisfy in practice, given that the dimension of $\mathbf{s}_t$ is much larger than that of $\mathbf{u}_t$. 

At the instant $t$, we first define the predicted SOI sequence 
as $\bar{\mathbf{x}}_t = [\mathbf{x}^\intercal_t, \mathbf{x}^\intercal_{t+1}, ..., \mathbf{x}^\intercal_{t+T-1} ]^\intercal $ and the control sequence as $\bar{\mathbf{u}}_t = [\mathbf{u}^\intercal_t, \mathbf{u}^\intercal_{t+1}, ..., \mathbf{u}^\intercal_{t+T-1}]^\intercal $.
Here $T$ is the prediction horizon. 
$\bar{\mathbf{x}}_t $ and $\bar{\mathbf{u}}_t$ are vectorized, resulting in the form of column vectors.
Following \prettyref{eq:mpc_opt}, the next subgoal $\mathbf{g}_{next}$ from $\mathcal{G}$ according to the current SOI $\mathbf{x}_t$ is determined.
Thus, the target SOI sequence is defined by $\bar{\mathbf{k}}_t = [\mathbf{g}_{next}^\intercal, \mathbf{g}_{next}^\intercal, ..., \mathbf{g}_{next}^\intercal]^\intercal $. 
Then, the cost function $\mathcal{J}$ can be formulated as:
\begin{align}\label{eq:mpc_cost_fn}
    \mathcal{J} = {\left( {{\bar{\mathbf{x}}_t} - {\bar{\mathbf{k}}_t}} \right)^\intercal} \mathbf{Q} \left( {{\bar{\mathbf{x}}_t} - {\bar{\mathbf{k}}_t}} \right) + \bar{\mathbf{u}}_t^\intercal \mathbf{R} {\bar{\mathbf{u}}_t},
\end{align}
where $\mathbf{Q}$ and $\mathbf{R}$ are weight matrices for state errors and control inputs, respectively.
State constraints $C_6$ on $\mathbf{x}_t$ from bag's physical property, control input constraints $C_7$ on $\mathbf{u}_t$ from dual-arm limits, and system modeling constraints $C_8$ from \prettyref{eq:sys_model}.
At each instant, MPC solves the following optimization problem over the prediction horizon $T$:
\begin{align}
\mathop{\arg\min}_{\bar{\mathbf{u}}_t}
\mathcal{J}
\ , \
\text{s.t.}~C_6,C_7,C_8.
\end{align}
The solution is an optimal control sequence $\bar{\mathbf{u}}^*_t = [{\mathbf{u}_t^*}^\intercal, {\mathbf{u}_{t+1}^*}^\intercal, ..., {\mathbf{u}_{t+T-1}^*}^\intercal]^\intercal \in \mathbb{R}^{12T}$.
Only the first control input from $\bar{\mathbf{u}}^*_t$ is applied to the system, i.e., ${\mathbf{u}_t^*}^\intercal$.
The system state evolves to $\mathbf{x}_{t+1}$ based on the applied control input and system model \prettyref{eq:sys_model}. After the dual-arm execution, the prediction horizon is shifted forward by one time step, and the above optimization process is repeated. This receding horizon approach ensures continuous adaptation to changes in the system or environment.


Due to the complicated property of the fabric bag, it is hard to obtain the analytical formulation of $\mathbf{J}_t$. 
Thus, we employ the Broyden approach \cite{broyden1965class} to approximate $\hat{\mathbf{J}}_{t}$ online. 
Specifically, it can be calculated by
$
\hat{\mathbf{J}}_{t} = \hat{\mathbf{J}}_{t-1} + \epsilon \cdot ({{ {{\mathbf{s}_t} - \hat{\mathbf{J}}_{t-1}{\mathbf{u}_t}}}}) \ / \ 
({{\mathbf{u}_t^\intercal {\mathbf{u}_t}}}) \cdot 
\mathbf{u}_t^\intercal,
$
with $\epsilon \in (0,1]$ controlling the convergence rate.
Afterwards, $\hat{\mathbf{J}}_{t}$ is used to replace ${\mathbf{J}}_{t}$.
Therefore, the state variable $\bar{\mathbf{x}}_t$ can be determined iteratively in the optimization process by following equations:
$
    \bar{\mathbf{x}}_t
    = \mathbf{A} \mathbf{x}_t + \mathbf{B} {\bar{\mathbf{u}}_t},
    \mathbf{A}  = \mathbf{I}_{T \times 1} \otimes \mathbf{E}_{3n_x},
    \mathbf{B} = \mathbf{L}_h \otimes \hat{\mathbf{J}}_t.    
$
Here $\mathbf{I}_{T \times 1}$ is the $T \times 1$ matrix of ones, and $\mathbf{E}_{3n_x}$ refers to the $3n_x \times 3n_x$ identity matrix.
$\mathbf{L}_h$ is the low triangle matrix of $\mathbf{I}_{T \times T}$, and $\otimes$ is the Kronecker product.

\commentTom{
This work uses shape servoing to manipulate the bag, with visual servoing as the core component. 
Visual servoing focuses on rigid objects, while shape servoing deals with deformable ones \cite{qi2023adaptive}. 
The employed MPC utilizes visual feedback to accomplish the bagging task. 
Specifically, $\bar{\mathbf{x}}_t$ represents the extracted visual feature points as defined in \eqref{eq:mpc_cost_fn}, while $\bar{\mathbf{k}}_t$ denotes the trajectories planned with CBiRRT.
}


\vspace{-15pt}
\section{Experiments}
\label{sec:exp}
\subsection{Experimental Setup}
\label{section7a}

\prettyref{fig:first_fig} shows the experimental setup. Two CR5 robotic arms are equipped with custom 3D-printed mounts to grasp each handle of the bag, secured with zip ties to prevent slippage. A D455 depth camera is fixed in an eye-to-hand configuration for top-down observation of the manipulation process at 640x480 resolution. Visual processing is done with OpenCV, generating point clouds $\mathcal{P}_t$ via RealSense libraries.
The velocity signal $\mathbf{u}_t$ is subject to strict limitations to ensure precision in estimating $\mathbf{J}_t$. 
Movement control is implemented in the ROS framework at about 11 Hz.
The 3D scanning devices (CR-Scan Ferret Pro) are used to acquire the vertex set $\mathcal{V}$ for each item $\mathcal{B}$. ArUco markers are attached to $\mathcal{B}$, enabling the robotic system to recognize the object category through its vision system before manipulation, facilitating access to the appropriate configuration associated with $\mathcal{V}$.

\vspace{-5pt}
\subsection{Experiments on GMM-based SOI Extraction}

\begin{figure}
    \centering
    \includegraphics[width=0.486\textwidth]{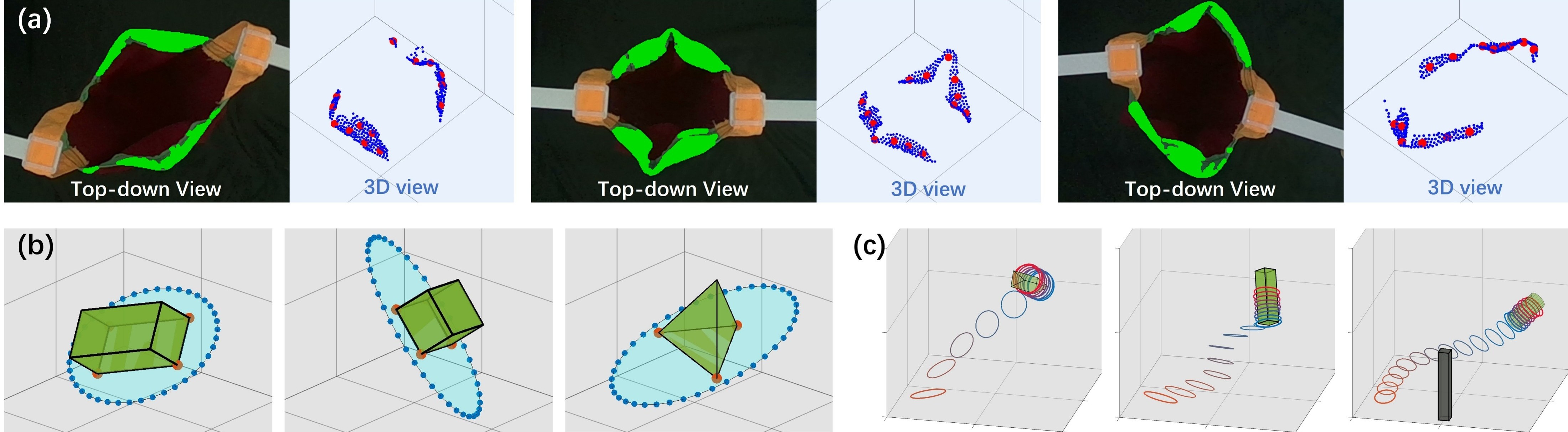}
    \vspace{-20pt}
    \caption{\commentBenji{(a) GMM-based SOI extraction results. Green area: highlighted bag rims. Blue points: point clouds $\mathcal{P}_t$. Red points: extracted SOI points $\mathcal{Q}_t$. (b) Generations of bagging SOI $\mathbf{g}_\dag$, as shown in ellipse dots (blue). The bottom of the baggable object $\mathcal{B}$ (green polyhedron) is represented by a set of vertices $\mathcal{V} = \{\mathbf{v}_i\}_{i = 1}^{n_v}$ (red dots). (c) Planned path $\mathcal{G}$ including $\mathcal{G}_\textit{pre-bagging}$ and $\mathcal{G}_\textit{bagging}$. The objects designated for bagging are highlighted in green, while the environmental obstacles are represented by black cuboids. }
    }
    \label{fig7}
    \vspace{-20pt}
\end{figure}

This section evaluates the GMM-driven SOI extraction from Sec. \ref{section4a}, which derives a clear SOI point set $\mathcal{Q}_t$ from the unfiltered, high-density, noise-affected point cloud $\mathcal{P}_t$.
The EM algorithm addresses \eqref{eq48}, allowing for a concise representation of $\mathcal{Q}_t$.
Fig. \ref{fig7}-(a) displays the outcomes produced by the GMM-based SOI extraction method, {where the rim of the bag is emphasized in green and the estimator successfully reconstructs a largely complete representation of the state $\mathcal{Q}_t$.}


\subsection{Experiments on Bagging SOI Generation}
This section evaluates the bagging SOI generation in Sec.~\ref{section4b}, which creates a pre-formed enclosure $\mathbf{g}_{\dag}$ around the bottom of $\mathcal{B}$.
\commentTom{
We conduct a total of 1,000 simulations on 12 different shapes, with several representative examples shown in Fig. \ref{fig7}-(b), each linked to a set of $\mathcal{V}$.
$\lambda_1$ constrains $\mathcal{V}$ of $\mathcal{B}$ within the generated ellipse. 
$\lambda_2$ positions the ellipse center close to $^m \bar{\mathbf{v}}$, enforcing overlap with $\mathcal{V}$. 
$\lambda_3$ regulates the parallelism between the ellipse and $\mathcal{V}$, maximizing coverage within the equal area ellipse.
Through offline simulations, these parameters are optimized to $\lambda_1 = 0.912, \lambda_2 = 0.007, \lambda_3 = 0.9943$.
The real perimeter of the bag's rim is measured as $R=0.68$m.
}


\commentTom{
The results show that effective bagging SOI $\mathbf{g}_{\dag}$ can be generated for various objects and placements. 
In each scenario, $\mathbf{g}_{\dag}$ meets the perimeter constraint $R$.
The enclosure of $\mathcal{V}$ validates $C_1$, while the central alignment of $\mathbf{g}_{\dag}$ demonstrates $C_2$'s efficacy. 
For non-uniform $\mathcal{V}$, $C_3$ ensures comprehensive coverage along the principal axis, especially for cuboids and triangular pyramids. 
$C_1$ and $C_2$ contribute to a uniform central distribution of $\mathbf{g}_{\dag}$ around $\mathcal{V}$, while $C_3$ ensures that the ellipse and $\mathcal{V}$ maintain the reasonable relative position by correcting the orientation. 
The average values for $C_1,C_2,C_3$ across 1000 simulations are 0.826, 0.005, and 0.9978, respectively. 
Simulations reveal that $C_1$ and $C_3$ regulate $\mathbf{g}_{\dag}$ more effectively than $C_2$ for the bagging SOI.
}




\vspace{-5pt}
\subsection{Experiments on SOI Planning}
This section evaluates the SOI planning strategy detailed in \prettyref{sec:soi_plan}, which constructs a collision-free deformation trajectory $\mathcal{G}$ that transitions smoothly from the initial $\mathbf{g}_0$ to the target $\mathbf{g}_*$, using the intermediate bagging SOI $\mathbf{g}_\dag$ as guidance. 
\commentTom{
$\lambda_4$ constrains the perimeter similarity between $\mathcal{X}_{rand}$ and the fitted counterparts.
$\lambda_5$ ensures concentricity, guaranteeing that the fitted shape closely resembles $\mathcal{X}_{rand}$ to maintain the shape uniqueness.
We empirically set $\lambda_4 = 0.002$ and $\lambda_5 = 0.021$ through simulations, respectively. 
}

Fig. \ref{fig7}-(c) shows six planning results $\mathcal{G}$ based on different configurations of $(\mathbf{g}_0, \mathbf{g}_\dag, \mathbf{g}_*)$. The gradient paths from orange to blue illustrate $\mathcal{G}_\textit{pre-bagging}$, while those from blue to red correspond to $\mathcal{G}_\textit{bagging}$. The ``Regularization'' ensures smoothness along $\mathcal{G}$ and compliance with the physical perimeter constraint $C_4$ (see \prettyref{eq:constraint_4}). The short transition between $\mathbf{g}_\dag$ and $\mathbf{g}_*$, along with their proximity to $\mathcal{B}$, results in noticeable variation in $\mathcal{G}_\textit{bagging}$.
Additionally, Fig. \ref{fig7}-(c) demonstrates that CBiRRT can produce a feasible deformation trajectory $\mathcal{G}$ in obstacle-rich environments, with the black cuboid representing a user-specified obstruction. The planning outcomes confirm that CBiRRT constructs a continuous path $\mathcal{G}$, while \prettyref{eq:constraint_4} ensures each node adheres to perimeter constraints. These results validate the optimization strategy in \prettyref{eq:project_stable}, showing that additional constraints can enhance planning precision and meet specific task objectives. Note that, in this paper, $\mathbf{g}_*$ is represented as a simple translation from $\mathbf{g}_\dag$, but it may involve a more complex configuration.

\begin{table*}[!tbp]
    \caption{Experiments Results (S.R.: success rate. Manip.: Manipulation)}
    \vspace{-10pt}
    \centering
    \resizebox{0.95\textwidth}{!}{%
    \begin{tabular}{cccccccccccccccc}
    \toprule
        \multirow[m]{3}{*}{Method}	& \multicolumn{3}{c}{Coffee box (Exp 1)}	& \multicolumn{3}{c}{Canned pineapple (Exp 2)} & \multicolumn{3}{c}{Grapefruit (Exp 3)} & \multicolumn{3}{c}{Triangular prism (Exp 4)} & \multicolumn{3}{c}{Tea caddy (Exp 5)}\\
        \cmidrule(lr){2 - 4}
        \cmidrule(lr){5 - 7}
        \cmidrule(lr){8 - 10}
        \cmidrule(lr){11 - 13}
        \cmidrule(lr){14 - 16}
        & Planning & Planning & Manip. & Planning & Planning & Manip. & Planning & Planning & Manip. & Planning & Planning & Manip. & Planning & Planning & Manip. \\
        & S.R. & time (s) & S.R. & S.R. & time (s) & S.R. & S.R. & time (s) & S.R. & S.R. & time (s) & S.R.  & S.R. & time (s) & S.R.\\
        \midrule
        FFG-RRT \cite{roussel2014motion}
        & 6/10
        & $3.87 \pm 1.97$
        & 8/8
        
        & 8/10
        & 2.37 $\pm$ 0.87 
        & 8/8
        
        & 7/10
        & 3.89 $\pm$ 1.18
        & 8/8
        
        & 6/10
        & 3.58 $\pm$ 1.11
        & 8/8
        & 8/10    & 4.93 $\pm$ 1.23   & 7/8 
        \\
        TS-RRT \cite{suh2011tangent}
        & 7/10
        & 6.32 $\pm$ 1.08 
        & 8/8
        
        & 8/10
        & 5.58 $\pm$  1.13
        & 8/8
        
        & 9/10
        & 6.85 $\pm$ 0.56 
        & 8/8
        
        & 7/10
        & 7.32 $\pm$ 1.34 
        & 8/8
        & 8/10   & 7.46 $\pm$ 0.97  & 8/8
        \\
        
        IBVS \cite{ren2020image}
        & -
        & -
        & 4/8
        
        & -
        & - 
        & 7/8
        
        & -
        & - 
        & 5/8
        
        & -
        & - 
        & 6/8
        & -  & -  & 7/8
        \\
        
        SSVS \cite{hao2011universal}
        & -
        & - 
        & 5/8
        
        & -
        & - 
        & 7/8
        
        & -
        & - 
        & 6/8
        
        & -
        & - 
        & 7/8
        & - & - & 8/8
        \\
        
        \textbf{Ours}
        & 9/10
        & 5.13 $\pm$ 1.26 
        & 8/8
        
        & 10/10
        & 4.21 $\pm$ 0.98 
        & 8/8
        
        & 10/10
        & 4.98 $\pm$ 1.93 
        & 8/8
        
        & 9/10
        & 5.32 $\pm$1.56 
        & 8/8 
        & 10/10  & 6.16 $\pm$ 1.58  & 8/8
        \\
        \midrule
        \multirow[m]{3}{*}{Method}	
        & \multicolumn{3}{c}{Work bin (Exp 6)}	
        & \multicolumn{3}{c}{Tea bucket (Exp 7)}
        & \multicolumn{3}{c}{Plastic packing (Exp 8)} 
        & \multicolumn{3}{c}{Tea bucket (Exp 9)}
        & \multicolumn{3}{c}{3D-printed cuboid (Exp 10)}
        \\
        \cmidrule(lr){2 - 4}
        \cmidrule(lr){5 - 7}
        \cmidrule(lr){8 - 10}
        \cmidrule(lr){11 - 13}
        \cmidrule(lr){14 - 16}
        & Planning & Planning & Manip.
        & Planning & Planning & Manip. 
        & Planning & Planning & Manip. 
        & Planning & Planning & Manip. 
        & Planning & Planning & Manip. 
        \\
        & S.R. & time (s) & S.R. 
        & S.R. & time (s) & S.R. 
        & S.R. & time (s) & S.R. 
        & S.R. & time (s) & S.R. 
        & S.R. & time (s) & S.R. 
        \\
        \midrule
        FFG-RRT \cite{roussel2014motion}
        & 7/10    & 3.88 $\pm$ 0.98   & 8/8
        & 8/10    & 3.93 $\pm$ 1.06   & 8/8
        & 7/10       & 3.67 $\pm$ 1.66    & 6/8        
        & 8/10       & 2.26 $\pm$  0.97   & 7/8
        & 7/10       & 3.76 $\pm$ 1.03    & 6/8
        \\
        TS-RRT \cite{suh2011tangent}
        & 8/10   & 6.13 $\pm$ 1.21  & 8/8
        & 9/10   & 5.98 $\pm$ 0.72  & 8/8        
        & 7/10  & 5.96 $\pm$ 1.23  & 7/8
        & 8/10  & 5.12 $\pm$ 0.98  & 7/8
        & 9/10  & 5.15 $\pm$ 0.83  & 7/8        
        \\
        
        IBVS \cite{ren2020image}
        & -  & -  & 7/8
        & - & -  &  7/8
        & -  & -  & 5/8
        & -  & -  & 5/8
        & - & -  & 6/8
        \\
        
        SSVS \cite{hao2011universal}
        & - & - & 7/8
        & - & - & 7/8
        & - & - & 6/8
        & - & - & 6/8
        & - & - & 6/8
        \\
        
        \textbf{Ours}
        & 9/10  & 4.92 $\pm$ 0.96  & 8/8
        & 10/10  & 5.06 $\pm$ 1.36  & 8/8
        & 8/10    & 5.68 $\pm$ 1.29  & 7/8
        & 9/10  & 4.06 $\pm$ 1.13    & 8/8
        & 10/10  & 5.31 $\pm$ 1.48  & 7/8
        \\
    \bottomrule
    \end{tabular}
    }
    \label{table2_new}
    \vspace{-10pt}
\end{table*}

\subsection{Dual-arm Bagging Manipulation}

Dual-arm bagging experiments assess the effectiveness of the proposed manipulation method. Four distinct baggable objects are used in Exp. 1-4: a coffee box, canned pineapple, a grapefruit, and a 3D-printed triangular prism. The procedure involves the dual-CR5 robotic system guiding the bag from different initial positions along the pre-bagging trajectory $\mathcal{G}_\textit{pre-bagging}$ to the intermediate configuration $\mathbf{g}_\dag$, followed by bagging path along $\mathcal{G}_\textit{bagging}$ to reach the final target shape $\mathbf{g}_*$ and complete the operation. To evaluate the performance, we compare two motion planning algorithms, FFG-RRT \cite{roussel2014motion} and TS-RRT \cite{suh2011tangent}, alongside two control strategies, IBVS \cite{ren2020image} and SSVS \cite{hao2011universal}.

\begin{figure}[!tbp]
\centering
\includegraphics[width=0.48\textwidth]{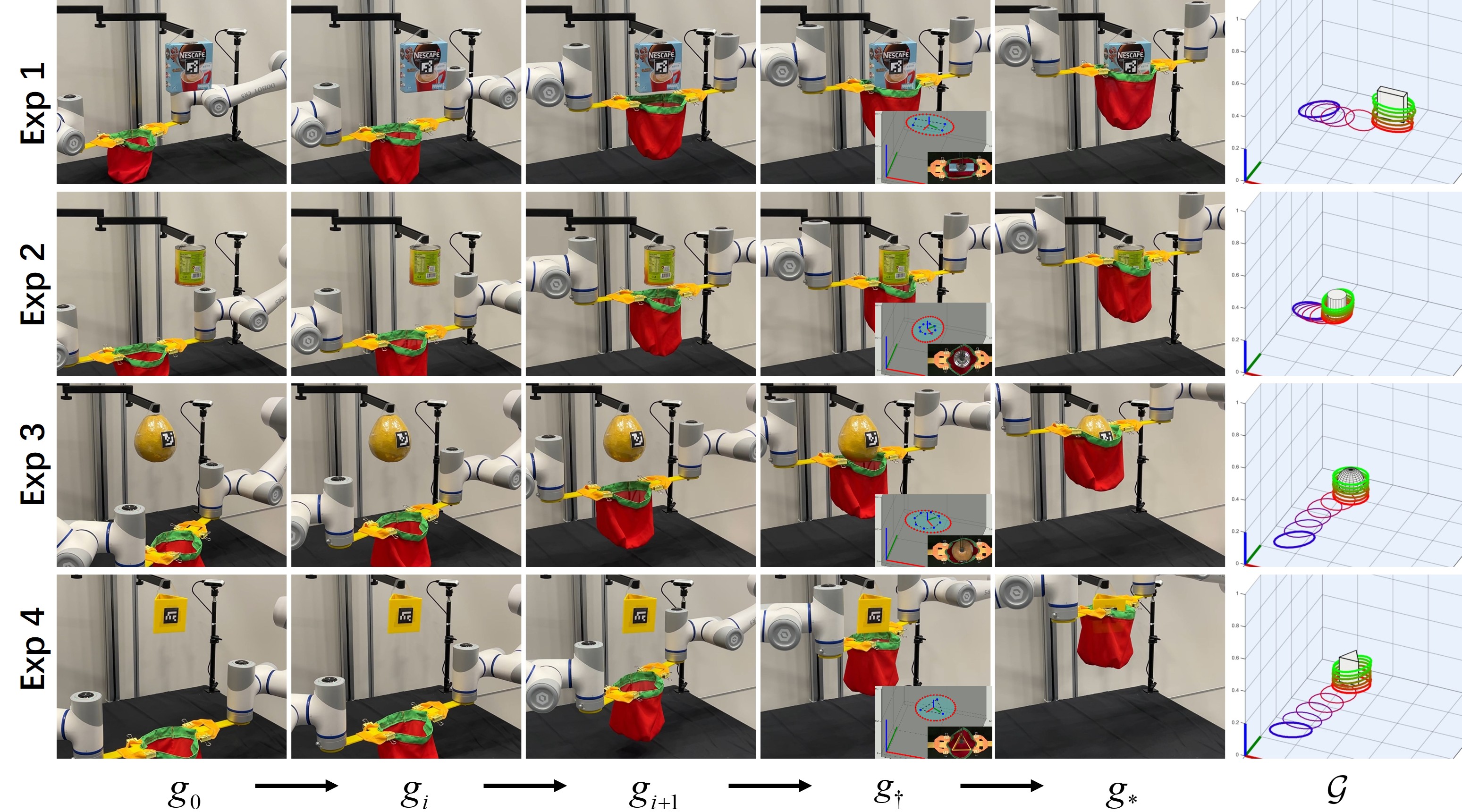}
\vspace{-20pt}
\caption{Dual-arm bagging manipulation. Insets in the 4-th column show the determined bagging SOIs. The last column displays the planned SOI trajectories. The animation of the bagging process can be seen in the video. 
}
\label{fig14}
\vspace{-20pt}
\end{figure}

%
Fig. \ref{fig14} shows the bagging process from Exp. 1-4. 
For a quantitative evaluation, we introduce three key metrics: planning success rate, planning duration, and manipulation success rate, reflecting the effectiveness of the planning and control algorithms. A detailed comparison is provided in Table~\ref{table2_new}.
The planning success rate indicates that CBiRRT outperforms other methods, achieving high reliability with reasonable computational cost. In contrast, FFG-RRT excels in speed, offering the shortest planning time due to its direct forward exploration strategy, while CBiRRT's bidirectional approach emphasizes stability, resulting in more search steps.
Regarding manipulation success, the MPC clearly delivers the best results, surpassing the other control strategies. This is mainly because conventional shape servoing assumes a static desired shape, while the bagging task requires tracking evolving deformation trajectories. MPC's predictive capabilities align well with this dynamic need, enhancing tracking stability. These findings highlight MPC's suitability for complex robotic manipulation tasks. Additionally, $\mathbf{g}_\dag$ serves as an intermediate buffer, segmenting the bagging process into pre-bagging and bagging phases, which enhances the robustness of the manipulation and contributes to a higher success rate.

\begin{figure*}
\centering
\includegraphics[width=0.98\textwidth]{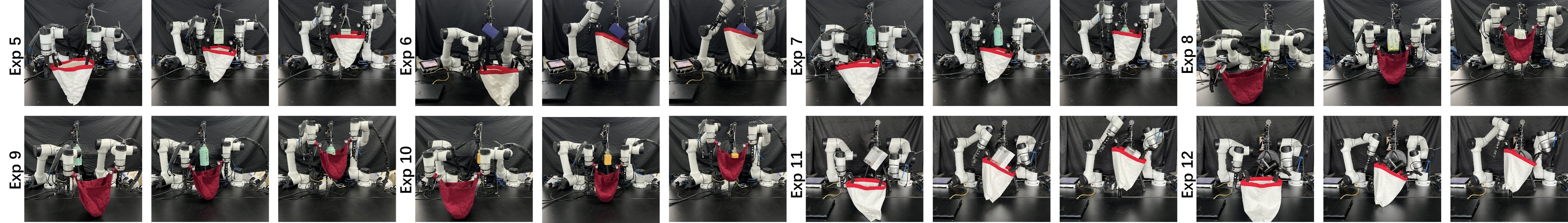}
\vspace{-10pt}
\caption{Dual-arm manipulation using a red-edged bag and a solid-color bag for single (Exp. 5-10) and bound (Exp. 11-12) objects.
}
\label{fig_add_exp}
\vspace{-20pt}
\end{figure*}

\subsection{\commentTom{Validations on Different Bags and Objects}}

To further validate the effectiveness of our method, we introduced additional bags and numerous objects with different poses. Here we used two bag types: one with color-contrasting red edges (Exp. 5-7) and another in solid color (Exp. 8-10).
The results in Fig. \ref{fig_add_exp} show that the proposed method remains effective in new task environments. Table \ref{table2_new} also provides a detailed comparison between our method and baselines. Exp. 5-7 shows a higher planning success rate than Exp. 8-10, likely due to larger bag dimensions. Notably, our method maintains the highest success rate among all approaches. In experiments with smaller objects, all three methods exhibited similar planning times, though our approach was slightly faster. 
The larger bags in Exp. 5-7 correspond to higher manipulation success rates, attributed to two factors: (1) the wider bag openings facilitate easier expansion, and (2) occlusion issues during manipulation for the solid-color bags stem from the neural network's detection accuracy limitations. 
\commentBenji{In addition, Exp. 11 and 12 demonstrate the bagging operations for bound objects. 
From Fig. \ref{fig_add_exp}, the dual-arm system precisely handles the bag into the bottom of the bound objects (2 objects bundled in Exp. 11 and 3 objects bundled in Exp. 12) and completely wraps around them.} 
\commentTom{
Notably, in Exp. 6, 11, and 12, the bag successfully envelops the tilted objects along their bottom axis.
}
In summary, our system has been validated to accurately and robustly complete various bagging tasks.




\subsection{\commentTom{Comparison with Baselines}}
\commentTom{
We replicate three dual-arm manipulation approaches, i.e., \emph{ShakingBot} \cite{gu2024shakingbot}, \emph{AutoBag} \cite{chen2023autobag},  and \emph{BimaManip.}\cite{zhou2024bimanual} for the task in Exp. 1. 
The first two use action primitives, while \emph{BimaManip.} and this study employ SOI-based shape servoing. 
Since the initial state of bags in our paper is already open, we have ignored the shaking operations originally employed in \cite{gu2024shakingbot} and \cite{chen2023autobag}.}
\commentTom{
Results are shown in Table \ref{table1}, where \textbf{Error} measures the distance between the SOI center and the object's center after task completion, indicating alignment accuracy.}
\commentTom{
From Table \ref{table1}, \emph{Ours} has the highest success rate and lowest error, indicating superior reliability and accuracy. 
Since \emph{ShakingBot} and \emph{AutoBag} use motion primitive-based action encoding for faster manipulation, \emph{AutoBag} achieves the shortest time. 
In contrast, this study focuses on tracking transition shapes from CBiRRT, resulting in the slowest speed but the lowest error. 
\emph{BimaManip.} offers balanced performance with moderate time and accuracy.
Overall, \emph{Ours} shows the most robust and precise bagging performance.
}



\vspace{-10pt}
\section{Conclusion}
\label{sec:conclusion}

In this paper, we present a robotic system for automating bagging operations using a novel constraint-aware SOI planning framework for soft fabric bags. 
A key innovation is the system's ability to estimate and target the SOI, allowing precise control over the bag's opening rim and enhancing operational efficiency.
Additionally, we introduce an adaptive visual servoing system and a robust framework that showcases the system's adaptability to environmental constraints. 

\vspace{-10pt}
\begin{table}[ht]
	\caption{\commentTom{Bagging comparison across multiple approaches.}}
    \vspace{-10pt}
	\label{table1}
	\centering
	\begin{tabular}{cccc}
    \hline
    \thead[c]{} & 
    \thead[c]{Manip. S.R.     } &
    \thead[c]{Manip. Time (s) } &
    \thead[c]{Error       (cm) }      \\
    \toprule
    ShakingBot \cite{gu2024shakingbot} 
    & 4/8   & 39.86 $\pm$ 3.74  & 6.47 $\pm$ 1.13     \\
    
    AutoBag \cite{chen2023autobag} 
    & 6/8   & 32.54 $\pm$ 4.11  & 5.36 $\pm$ 0.87     \\
    
    BimaManip. \cite{zhou2024bimanual} 
    & 7/8   & 44.21 $\pm$ 5.81  & 3.50 $\pm$ 0.93     \\
    
    \textbf{Ours}
    & 8/8   & 53.09 $\pm$ 2.92  & 2.10 $\pm$ 0.77     \\
    \bottomrule     
    \end{tabular}
    \vspace{-15pt}
\end{table}

\bibliographystyle{IEEEtran}
\typeout{}
\bibliography{IEEEabrv,mybibfiles}
\end{document}